\documentclass{article}
\usepackage{amsfonts}
\usepackage{spconf,amsmath,graphicx,epsfig}
\usepackage{subcaption}
\usepackage{mwe}
\usepackage{xcolor}
\usepackage{hyperref}
\usepackage{algorithm}
\usepackage{algorithmic}
\usepackage{multirow}

\title{Personalization strategies for end-to-end speech recognition systems}
\name{%
\begin{tabular}{@{}c@{}}
Aditya Gourav$^{\ast}$ \qquad
Linda Liu$\sthanks{Equal contribution}$ \qquad
Ankur Gandhe \qquad
Yile Gu  \\
Guitang Lan \qquad
Xiangyang Huang \qquad
Shashank Kalmane  \qquad
Gautam Tiwari  \\
Denis Filimonov \qquad
Ariya Rastrow \qquad
Andreas Stolcke \qquad
Ivan Bulyko
\end{tabular}}

\address{Amazon Alexa}

\begin{document}
\ninept
\maketitle
\begin{abstract}
The recognition of personalized content, such as contact names,
    remains a challenging problem for end-to-end speech recognition systems.
    In this work, we demonstrate how first- and second-pass rescoring
    strategies can be leveraged together to improve the recognition of such words. 
    Following previous work, we use a shallow fusion approach to bias towards recognition of personalized content in the
    first-pass decoding. We show that such an
    approach can improve personalized content recognition by up to 16\% with minimum degradation on the general use
    case.
    We describe a fast and scalable algorithm that enables our biasing models to remain at the word-level, while
    applying the biasing at the subword level. This has the advantage of not requiring the biasing models to 
    be dependent on any subword symbol table.
    We also describe a novel second-pass de-biasing approach: used in conjunction with a first-pass shallow fusion
    that optimizes on oracle WER, we can achieve an additional 14\% improvement on personalized content recognition, and
    even improve accuracy for the general use case by up to 2.5\%.
\end{abstract}
\section{Introduction}
\label{sec:intro}

The successful recognition of personalized content, such as a user's contacts or custom smart home device names,
is essential for automatic speech recognition (ASR). Personalized content recognition is challenging as such words can 
be very rare or have low probability for the user population overall. For instance, a user's contact list may contain foreign names or
unique nicknames, and they may freely name their smart home devices.

This problem is exacerbated for end-to-end (E2E) systems, such as those based on CTC \cite{graves2012sequence}, LAS
\cite{chan2015listen}, or RNN-T \cite{graves2014towards}. Unlike hybrid ASR systems, 
which include acoustic and language model (LM) components that are trained separately, E2E systems use a single network that is trained end-to-end. Whereas in a hybrid system, the LM component can be trained separately on any written text, in an E2E system, the training is generally restricted to acoustic-text pairs. 
As a result, E2E systems are often trained with less data than hybrid systems, making personalized content
recognition particularly challenging given the limited representation during training. Furthermore, hybrid systems are able to incorporate 
personal content into the decoding search graph, i.e., via class-based language models and on-the-fly composition of
biasing phrases and n-grams \cite{aleksic2015improved, brown1992class, mcgraw2016personalized, hall2015composition, williams2018contextual}.

Various approaches have been proposed for improving personalized content recognition for E2E models, including model
fine-tuning with real or synthesized audio data \cite{sim2019personalization}, incorporating personalized content
directly during E2E training using a separate bias-encoder module \cite{pundak2018deep}, using a token passing decoder
with efficient token recombination during inference \cite{chen2019end}, and shallow fusion
(e.g., \cite{toshniwal2018comparison, williams2018contextual, zhao2019shallow, huang2020class}).

In this work, we describe a novel approach to address this problem using a combination of first-pass shallow fusion and
second-pass rescoring. We first provide a comparison of a few shallow fusion approaches: shallow fusion applied at the word-level and
subword level, as well as contextual shallow fusion. We describe a novel algorithm that uses grapheme-level
lookahead to perform subword-level rescoring, thus bypassing the need to build subword-level language models that are
dependent on the wordpiece model that generates the subwords.
We show the benefit of contextual shallow fusion in capturing
improvement in personalized content recognition.
Finally, we describe a novel de-biasing approach in which we treat the second-pass rescoring as an optimization
problem to optimally combine scores from the E2E model, the shallow fusion model, and second-pass LMs. Apart from improving recognition for the personalized content, it also improves the general recognition. 

\section{Previous work}
\label{sec:prev}

One popular approach to improve personalized content recognition is via shallow fusion \cite{kannan2018analysis}. In
shallow fusion, the scores from an external language model $Score_{SF}(y)$ scaled by a factor $\lambda$ are combined
with the main decoding scores $P_{RNNT}(y \mid x)$ during beam search:

\begin{equation}
    \hat{y}=\arg \max_{y} (\log P_{RNNT}(y \mid x)+\lambda \log Score_{SF}(y))
\label{eq:sf}
\end{equation}

This biasing can be applied at word boundaries \cite{williams2018contextual}, at the grapheme level \cite{chen2019end,
zhao2019shallow, pundak2018deep}, or at the subword level \cite{zhao2019shallow, huang2020class}. 
Given that E2E models generally used a constrained beam \cite{li2019improving}, applying biasing only at word boundaries cannot improve
performance if the relevant word does not already appear in the beam. As a result, compared to grapheme-level biasing
which tends to keep the relevant words on the beam, word-level biasing results in less
improvement on proper nouns such as contact names \cite{pundak2018deep}. Applying biasing at the subword level, which would
result in sparser matches at each step of the beam compared to the grapheme level, results in further improvements
\cite{zhao2019shallow}.
%\cite{pundak2018deep} compared this
%approach of applying the bias at the end of the word, with the approaches of applying the bias at the first grapheme of
%the word and applying a weight at each grapheme unit of the word. When biasing at each grapheme unit, they also included an opposite subtractive
%cost, in order to prevent biasing toward words that do not match completely. They found that the latter method resulted in the best
%WER improvement and was more likely to keep the word on the beam. 

One challenge in applying biasing at the subword level, particularly for personalization, is that each of the biasing
models needs to be built at the subword level and include all possible segmentations of a given word. This can be expensive when we have one or more models per user,
particularly if
the wordpiece model used to train the first-pass model often changes. Unlike previous work, which generally relies on
composition with a speller FST to transduce a sequence of wordpieces into the corresponding word (e.g.,
\cite{zhao2019shallow, pundak2018deep, he2019streaming}), we
describe a novel prefix-matching algorithm in Section \ref{sec:alg} that enables the language models to be kept at the word level and applies the subword decomposition 
at inference time. 

%Another challenge of shallow fusion is that including too many biasing phrases can degrade performance on both the
%biased content and the general use case \cite{williams2018contextual}. Following previous work, we use a contextual
%biasing model in order to address this problem. We provide comparisons with various personalized models, both with and
%without this contextual model. We describe a practical solution to address the problem of degradation on the general use case.

Another challenge these shallow fusion approaches to personalization is how to improve recognition of
personalized content while not degrading performance on general non-personalized content; to this end, several
strategies for applying contextual biasing have been proposed \cite{zhao2019shallow, huang2020class, chen2019end}.
Many of these strategies reveal that applying shallow fusion in context minimizes, but does not completely remove, the
degradation observed on general data, and do not discuss the potential impact of second-pass rescoring.
For example, \cite{zhao2019shallow} finds that applying contextual shallow fusion decreases the negative impact
on general content while maintaining performance on the shallow fusion content; however, even in this best case, 
they report a slight degradation of 5.8\% (6.9 to 7.3 WER, cf. 12.5 for non-contextual shallow fusion) on general data. 

On one hand, a more aggressive shallow fusion model enables more personalized content to appear in the n-best
hypotheses but on the
other hand, it is also more likely to cause false recognitions of the biased personalized
content. To address this, we present a strategy in which we optimize shallow fusion for the n-best, as opposed to the
1-best, hypotheses, thereby maximizing the personalized content present in the n-best. To recover the correct 1-best, we
explore a novel second-pass de-biasing
approach that optimizes the combination of the E2E, shallow fusion, and second-pass scores.

\section{Methods}
\label{sec:exp}

\subsection{Baseline RNN-T model}
%R9 
Following \cite{guo2020efficient}, our baseline RNN-T model consists of an encoder comprised of five LSTM layers of size
1024, and a two-layer LSTM prediction network of
size 1024 with an embedding layer of 512 units. The softmax layer consists of 4k (subword) output units.
Our model was trained on over 200k hours of anonymized utterances from interactions with a voice assistant according to the minimum word error rate criterion \cite{shannon2017optimizing,
guo2020efficient}.

\subsection{First-pass shallow fusion}
\label{sec:fpsf}
\subsubsection{Personalized models}
For each anonymized user in our test set, we construct three personalized models, corresponding to (1) contact names
(2) smart home device names and (3) enabled application names. Each of these models is represented as a word-level weighted
finite state transducer (FST). An example is shown in Figure \ref{fig:rescoring_example}a. For simplicity, in our experiments, 
each word level arc has the same weight of -1. On average, each user has 600 personalized contact names, 50 device names, and
70 enabled applications.
%3503944/6016  = 580
%2582330
%416604/

\subsubsection{Subword rescoring with lookahead}
\label{sec:alg}

\begin{figure*}[ht!]
\begin{center}
\vspace{-1mm}
    \includegraphics[width=0.8\linewidth]{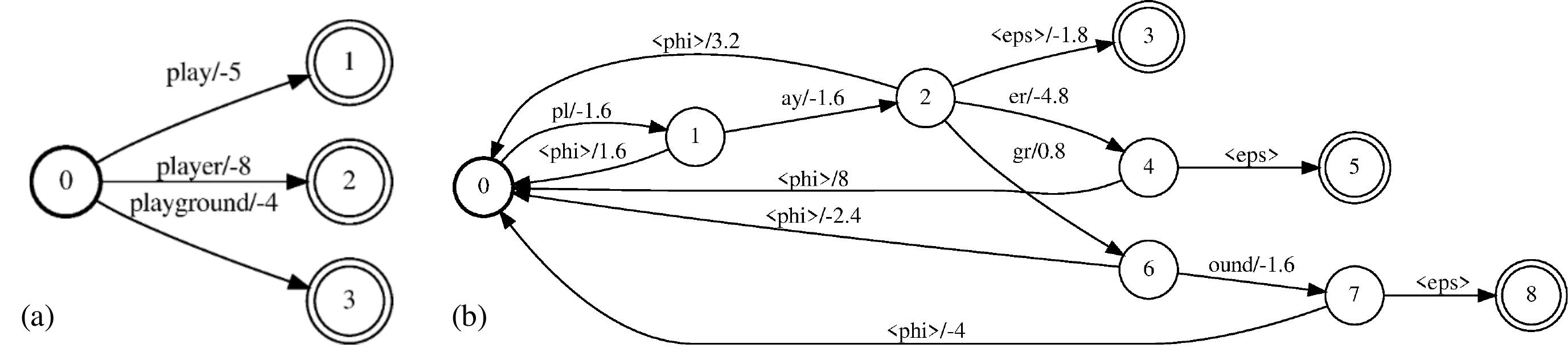}
\end{center} 
\vspace{-0.2cm}
    \caption{A illustration of (a) word-level biasing FST (b) subword-level FST with transition state weights evaluated
    using a grapheme-level lookahead. Additional phi self loop at the start state is not shown.} 
\label{fig:rescoring_example}
\end{figure*}

We describe our approach to biasing at the subword level using our word-level personalized models (Algorithm \ref{alg:subword_boost}). We leverage ideas similar to \cite{mohri_weight_pushing} for subword level lookahead weight pushing and start with a
word-level model represented as an FST, such as the one shown in
Figure \ref{fig:rescoring_example}a. In this case, there are three paths associated with this FST, containing the words
``play'', ``player'', and ``playground''. In Figure \ref{fig:rescoring_example}b, we show the subword breakdown for
these words, based on some wordpiece model. The weights on each path are determined via Algorithm
\ref{alg:subword_boost}. Notice that the net weight for each path remains the same: i.e., 
the weight between state 0 and 5 (representing the word ``player'') in the subword-level FST is (-1.6) +(-1.6)+(-4.8) =
-8, which is the same as the weight for the same word in the word-level FST. The weight $w_{pushed}$ for each transition state is
determined as follows: $w_{pushed} = (\frac{L}{N})(w_{lookahead})$, where L is the length of the prefix so far and N is
the longest length of all matched words. In our example, given the input sequence ``play'' (pl, ay, \textunderscore)
from Figure \ref{fig:rescoring_example}, we can see there are three arcs prefixed with the subword ``pl'': thus, we
have $L = 2$, $N=10$, and the pushed weight is  -8 * 2 / 10 = -1.6. Additionally, similar to \cite{pundak2018deep, he2019streaming}, we
add fallback arcs for each non-final state with a weight equal to the negation of the current total weight up to that
point.

This approach is beneficial as it avoids unnecessary arc expansion and provides a heuristic approach to perform
subword-level rescoring without the need to build the biasing FST itself directly at the subword level.
Additionally, this prefix matching approach enables us to consider any possible subword sequences for a word.
To optimize the search for arcs that have a common prefix string, we sort the input arc in lexicographic order so that 
we can use binary search to find the lower and upper bound of arc indices. As we continue to process subword input, we
are able to narrow down the search range quickly. We also cache all newly created states in
subword level FST $S$, which results in efficient weight evaluation. 

\begin{algorithm}[tb]
\begin{algorithmic}
%   \STATE {\bfseries 1.}  \textbf{Input}: $\alpha_{LM}$,$\alpha_{ID}$, $\alpha_{SF}$ \\
  \STATE $\mathbf{Expand}(T, s, i, W)$:
	\STATE {\bfseries 1.} Initialize : $R \leftarrow \phi$; $prefix \leftarrow \epsilon$; $w_{prev} \leftarrow 0$ \\
	\STATE {\bfseries 2.} Sort : $E_s$ by $i[E_s]$ in lexicographical order \\
\STATE {\bfseries 3.} \textbf{for} $sw$ in $W$ \textbf{do} \\
\STATE {\bfseries 4.} \hspace{1mm} \textbf{if} $sw$ is $delimiter$ \textbf{then}\\
\STATE {\bfseries 5.} \hspace{6mm}    \textbf{if} $prefix \in i[E_s]$ \textbf{then}\\
\STATE {\bfseries 6.} \hspace{10mm}    Return $R \cup (w_{e} - w_{prev}$)\\
\STATE {\bfseries 7.} \hspace{6mm}    \textbf{else}\\
\STATE {\bfseries 8.} \hspace{10mm}  Return  $R \cup w^{-1}(\pi(i, t))$ \\
\STATE {\bfseries 9.} \hspace{3mm} $prefix \leftarrow \mathbf{Concatenate}(prefix, sw)$ \\
\STATE {\bfseries 10.} \hspace{1mm} $E_s \leftarrow \textbf{BinarySearch}(prefix \in \mathbf{Prefix}(i[E_s]))$ \\
\STATE {\bfseries 11.} \hspace{1mm} \textbf{if} $E_s$ is empty \textbf{then} \\
\STATE {\bfseries 12.} \hspace{5mm} Return $R \cup w^{-1}(\pi(i, t))$ \\
\STATE {\bfseries 13.} \hspace{1mm} \textbf{else} \\
\STATE {\bfseries 14.} \hspace{5mm} $N \leftarrow$ max string length in $i[E_s]$\\
\STATE {\bfseries 15.} \hspace{5mm} $L \leftarrow$ prefix string length\\
\STATE {\bfseries 16.} \hspace{5mm} $w_{lookahead} \leftarrow \bigoplus w(e \in E_s))$ \\
\STATE {\bfseries 17.} \hspace{5mm} $w_{pushed} \leftarrow w_{lookahead} \cdot L / N $ \\
\STATE {\bfseries 18.} \hspace{5mm} Append:$R \leftarrow R \cup (w_{pushed} - w_{prev}$) \\
\STATE {\bfseries 19.} \hspace{5mm} $w_{prev} \leftarrow w_{pushed}$ \\
\STATE {\bfseries 20.} Return $R$
\end{algorithmic}
    \caption{On-the-fly subword rescoring with lookahead. $T$ and $s$ represent word level FST and a non-final state of it.
	$i$ denotes the starting state of subword level FST and i[e] the input symbol string for a transition $e$.
	$W$ is a sequence of subword input.  $R$ denotes a set of weights.
	$E_s$ denotes all transitions starting from $s$ in $T$.
	$\pi(a,b)$ represents a path from $a$ to $b$. $w_e$ is the net weight for transition $e$.
	$t$ is the previous state in subword FST.}
    \label{alg:subword_boost}
\end{algorithm}

\subsubsection{Contextual boosting model}
Following previous work such as \cite{aleksic2015improved, chen2019end, brown1992class, huang2020class}, we construct a class-based
language model containing three classes: contact names, home automation device names, and application names. To build the contextual biasing LM, we identified all utterances containing words that were annotated with aforementioned classes. We then replaced the word(s) in the utterance with the corresponding class tag (e.g., @contactname). All utterances with the replaced class tags that occurred a minimum of 10 times were included in the contextual biasing FST.
Unlike a typical class-based model, all arcs on the class-based model are unweighted. 
Weights only appear in the corresponding personalized models, which are injected at each class tag. Both the
class-based LM and personalized models operate at the subword level using the algorithm described in Section
\ref{sec:alg}.

\subsection{Datasets}
We evaluate on (1) a 20k utterance contact name test set and (2) a 20k utterance test set representing the general use
case. Both test sets consist of anonymized data from real user interactions with a personal assistant device.

\subsection{Second-pass rescoring}
We rescore 8-best hypotheses from the first-pass shallow fusion as described in Section \ref{sec:fpsf}. Each n-best hypothesis $y_{i}$ can be assigned a score based on the following equation:

\begin{equation}
\begin{aligned}
	Score(y_{i})= \log P_{RNNT}(y_{i} \mid x)+\alpha Score_{SF}(y_{i})  \\
	+ \beta \log P_{RLM}(y_{i})
\end{aligned}
\label{eq:2p}
\end{equation}

$P_{RLM}(y_{i})$ is the probability of the hypothesis $y_{i}$ assigned by the rescoring LM, $Score_{SF}(y_{i})$ is the shallow fusion score of $y_{i}$ from the first-pass. $\alpha$ and $\beta$ are the tunable scaling factors. In the tuning stage, we resort to a simulated annealing algorithm as described in \cite{xiangGenSA} to find the optimal values of $\alpha$ and $\beta$. The objective of the optimization is to minimize the overall WER of the dev set. This approach enables us to optimally combine multiple rescoring LMs with the first-pass scores.

To select the rescoring LM, we use the domain aware rescoring framework described in \cite{OURWORK} to differentiate
between the utterances with contact names and generic ones. For the generic utterances, we use an NCE based neural
LM (NLM) \cite{raju2019scalable} trained on 80 million utterances from live traffic. The model consists of
two LSTM layers, each with 512 hidden units. For the utterances with contact names, we use a KN-smoothed 4-gram class
based LM \cite{kn}, with a single ContactName class, trained on utterances with word annotations.

\section{Results and Discussion}
 
We report on word error rate reduction (WERR) and oracle WERR to the baseline RNN-T model throughout. The oracle WERR is
computed by finding the hypothesis in the 8-best that minimizes WER for each utterance.

\begin{table}[]
  \centering
    \begin{tabular}{lrrrr}
        \hline
        & \multicolumn{2}{c}{Contacts} & \multicolumn{2}{c}{General}\\
        Model  & WERR & Oracle & WERR & Oracle\\
        \hline
        RNN-T & -- & -- & -- & -- \\
        \hline
        \hspace{1mm} +word(1.0) & -6.5 & -2.3 & 0.4 & -0.2\\
        \hspace{1mm} +word(1.5) & -6.5 & -2.5 & 1.4 & 0.1\\
        \hspace{1mm} +word(2.0) & -5.0 & -1.6 & 3.0 & 0.1 \\
        \hline
        %subword=1 class
        \hspace{1mm} +noctxt-subwd(1.0) & -13.3 & -10.9 & -0.5 & 0.0\\
        \hspace{1mm} +noctxt-subwd(1.5) & -14.2 & -14.4 & 0.1 & 0.0\\
        \hspace{1mm} +noctxt-subwd(2.0) & -12.7 & -16.9 & 2.0 & 0.6 \\
        \hspace{1mm} +noctxt-subwd(2.5) & -7.8 & -18.4 & 5.0 & 0.8 \\
        \hspace{1mm} +noctxt-subwd(3.0) & -0.4 & -18.1 & 9.6 & 1.7 \\
        \hline
        %class (1) with unigram
        \hspace{1mm} +ctxt-subwd(1.0) & -14.0 & -10.9 & -0.5 & 0.0\\
        \hspace{1mm} +ctxt-subwd(1.5) & -16.3 & -13.6 & -0.3 & 0.0\\
        \hspace{1mm} +ctxt-subwd(2.0) & -16.5 & -16.9 & 0.6 & 0.3 \\
        \hspace{1mm} +ctxt-subwd(2.5) & -14.3 & -16.9 & 2.7 & 1.1 \\
        \hspace{1mm} +ctxt-subwd(3.0) & -10.8 & -16.7 & 5.0 & 2.5 \\
        \hline
    \end{tabular}
    \caption{Results using only the contact names personalized model, with different biasing weights, and comparing
    word-level biasing(word), with subword-level biasing with (ctxt-subwd) and without context(noctxt-subwd)}
    \label{table:word_subwrod_oneclass}
\end{table}

\subsection{Comparing shallow fusion approaches}
In Table \ref{table:word_subwrod_oneclass}, we report results comparing word-level biasing, to subword-level biasing
with and without context, using different biasing weights. We report results using only the personalized contact names model for shallow fusion. We find
significant improvements in WERR and oracle WERR when applying biasing at the subword level (best WERR improvement:
14.2\%) compared to the word
level (best WERR improvement: 6.5\%). This aligns with previous work \cite{pundak2018deep, zhao2019shallow}, which found that applying biasing at
the subword level allows more critical words to stay on the beam.

Comparing the subword results with and without context, we observe larger improvements in WERR on contact names at higher biasing
weights when using the contextual biasing model. For example, at a weight of 2.5, we observe improvements of 14.3\% with context, but only 7.8\% without
context. Additionally, we observe that constraining shallow fusion with context 
decreases the impact on the general WERR at higher biasing weights.

Finally, we observe that increasing the biasing
weight leads to improvements in oracle WERR on contact names, \emph{even when overall WERR improvements decrease}. This
suggests that a higher weight allows for more personalized content to appear in the n-best hypotheses,
even as it increases the number of false recognitions in the 1-best hypothesis. We return to this point later.

\subsection{Adding additional personalized content in shallow fusion}

In Table \ref{table:inc_personal_models}, we show that biasing in context helps to avoid degradation on general use
cases particularly as the number of classes increases. For these results, we use additional personal models (devices, applications).
We can observe that degradation on the general test set is more pronounced when the amount of biasing content 
increases. 
This is in line with previous work (e.g., \cite{williams2018contextual}).
Specifically, using a biasing weight of 2.5, we observe an 8.5\% degradation on the general test set without
context, but only 2.6\% with context. Critically, we observe that the WERR for contact names is
preserved.

\begin{table}[]
  \centering
    \begin{tabular}{lrrrr}
        \hline
        & \multicolumn{2}{c}{Contacts} & \multicolumn{2}{c}{General}\\
        Model  & WERR & Oracle & WERR & Oracle\\
        \hline
        RNN-T & -- & -- & -- & -- \\
        \hline
        %subword=3 class
        \hspace{1mm} +noctxt-subwd(1.0) & -14.2 & -11.6 & 0.1 & 0.0\\
        \hspace{1mm} +noctxt-subwd(1.5) & -13.1 & -15.3 & 2.6 & 0.5\\
        \hspace{1mm} +noctxt-subwd(2.0) & -8.8 & -17.3 & 8.5 & 1.7 \\
        \hspace{1mm} +noctxt-subwd(2.5) & 2.9 & -18.4 & 18.7 & 3.4 \\
        %maybe more lines
        \hline
        %class (3) with unigram
        \hspace{1mm} +ctxt-subwd(1.0) & -14.3 & -11.1 & -0.5 & 0.0\\
        \hspace{1mm} +ctxt-subwd(1.5) & -16.5 & -13.4 & 0.5 & 0.6\\
        \hspace{1mm} +ctxt-subwd(2.0) & -16.5 & -16.1 & 2.6 & 1.1 \\
        \hspace{1mm} +ctxt-subwd(2.5) & -14.2 & -16.5 & 5.7 & 2.5 \\
        \hspace{1mm} +ctxt-subwd(3.0) & -14.3 & -16.7 & 5.7 & 2.5 \\
        \hline
    \end{tabular}
    \caption{Results using three personalized models, with different biasing weights. Biasing with context 
    helps to avoid general degradation at the same level of biasing weights}
    \label{table:inc_personal_models}
\end{table}

\subsection{Second Pass Rescoring}
A trend seen in Table \ref{table:word_subwrod_oneclass} is that the 1-best WERR for both the Contacts and the
General test sets degrades as the shallow fusion biasing factor increases. However, Oracle WERR for the
Contacts test set improves. We address this divergence using second-pass rescoring.

We note that second-pass rescoring {\it without} shallow fusion provides an improvement of 16.5\% and 2.3\% on the
Contacts and General test sets. Following sections elicit that we see an additional 10-15\% improvement on the Contacts
test set when shallow fusion is used along with second-pass rescoring.  To the best of our knowledge, no previous work has shown this synergy.

\begin{table}[]
  \centering
    \begin{tabular}{lrrrr}
        \hline
        & \multicolumn{2}{c}{2P no de-biasing} & \multicolumn{2}{c}{2P w/ de-biasing}\\
        Model  & Contacts & General& Contacts & General \\
        \hline
        %subword=3 class
        +noctxt-subwd(2.5) & -21.2 & 0.6  & -28.5 & -2.7 \\
        +ctxt-subwd(2.0) & -25.3 & -1.7  & -27.4 & -2.5 \\
        %maybe more lines
        \hline
    \end{tabular}
    \caption{Results of de-biasing the shallow fusion scores for contact name personalized model in second-pass(2P). }
    \label{table:de-biasing}
\end{table}

\subsubsection{De-biasing shallow fusion scores}
We observe that re-weighting the shallow fusion scaled scores from the first-pass helps us achieve better WERR compared to adding it with the first-pass RNN-T scores. i.e., setting $\alpha=1$ during optimization in Equation \ref{eq:2p}. It helps in achieving better WERR for shallow fusion with or without context, as can be seen in Table \ref{table:de-biasing}. We call this method de-biasing in second-pass and use it in the results reported in the subsequent sections. 

We observe that de-biasing is especially useful when there is no context-based biasing in the first pass. It improves
recognition for personalized content without compromising the WER of the general test set.

\subsubsection{Adding additional personalized models in shallow fusion}
Second-pass optimization can not only recover from degradation in WERR but can also improve WERR when additional
personalized models are added to first-pass shallow fusion without context (Table
\ref{table:second_pass_sf_more_classes_at_no_context_2.5}). The first-pass degradation can be seen in Table
\ref{table:inc_personal_models} and is reproduced in Table \ref{table:second_pass_sf_more_classes_at_no_context_2.5}: we
observe that in general, incorporating more biasing models without context results in larger degradations on the
general test set. However, second-pass rescoring with de-biasing
enables us to completely recover from these degradations, while continuing to improve overall
contact name WERR. This aligns with our reasoning that second-pass can improve the
first-pass 1-best degradation as long the first-pass oracle WERR continues to improve.

In Table \ref{table:second_pass_sf}, we report the WERR post second-pass rescoring for various weights of
shallow fusion biasing, with and without context. 
As the biasing weight increases, we improve WERR for both the Contacts and General test sets.

\begin{table}[]
  \centering
    \begin{tabular}{lrrrr}
        \hline
        & \multicolumn{2}{c}{First-pass} & \multicolumn{2}{c}{2P w/ de-biasing}\\
        Model  & Contacts & General& Contacts & General \\
        \hline
        One biasing model \\
        +noctxt-subwd(2.5) & -7.8 & 5.0 &  -28.5 & -2.7 \\
        \hline
        Three biasing models \\
        +noctxt-subwd(2.5) & 2.9 & 18.4 & -29.1 & -2.7 \\

        \hline
    \end{tabular}
    \caption{Results of second-pass rescoring when more personalized models are added in shallow fusion}
    \label{table:second_pass_sf_more_classes_at_no_context_2.5}
\end{table}

\begin{table}[]
  \centering
    \begin{tabular}{lrrrr}
        \hline
        & \multicolumn{1}{c}{Contacts} & \multicolumn{1}{c}{General}\\
        Model  & WERR & WERR\\
        \hline
        %subword=3 class
        +noctxt-subwd(2.0) & -27.2 & -2.3 \\
        +noctxt-subwd(2.5) & -28.5 & -2.7 \\
        +noctxt-subwd(3.0) & -29.2 & -2.9 \\
        +noctxt-subwd(3.5) & -29.3 & -2.5 \\
        %maybe more lines
        \hline
        %class (3) with unigram
         +ctxt-subwd(2.0) & -27.4 & -2.5  \\
         +ctxt-subwd(2.5) & -28.4 & -2.2  \\
         +ctxt-subwd(3.0) & -30.0 & -2.3  \\
        \hline
    \end{tabular}
    \caption{Results of second-pass rescoring over the contact names personalized model, with different biasing weights, with and without context}
    \label{table:second_pass_sf}
\end{table}

\section{Conclusion}
In this work, we have presented several strategies to improve personal content recognition for end-to-end speech recognition systems. We have outlined a novel algorithm for efficient biasing of personalized content on the subword level at inference time. This helps us improve on personal content recognition by 14\% - 16\% compared to RNN-T. We also describe a novel second-pass optimization to improve recognition by an additional 13\% - 15\% without degrading the general use case. Combining the two strategies, we achieve 27\% - 30\% improvement overall in personal content recognition and about 2.5\% improvement on the general test set. We also elucidate ways to tackle degradation on the general test set when biasing the RNN-T model in the absence of any context.

\clearpage
\bibliographystyle{IEEEbib}
\bibliography{refs}

\end{document}